\pdfoutput=1
\documentclass[11pt, a4paper, logo, copyright, nonumbering]{hku}
\usepackage{natbib}

\usepackage{fancyhdr}
\usepackage[utf8]{inputenc} 
\usepackage[T1]{fontenc}    
\usepackage{booktabs}       
\usepackage{amsfonts}       
\usepackage{nicefrac}       
\usepackage{microtype}      
\usepackage{xcolor}         
\usepackage{float}

\usepackage{graphicx}
\usepackage{multirow}
\usepackage{enumitem}
\usepackage{subcaption}
\usepackage{xspace}
\usepackage{makecell}
\usepackage{tcolorbox}

\usepackage{fontawesome5}  
\usepackage{array}
\definecolor{mydarkblue}{rgb}{0,0.08,0.45}
\definecolor{linkcolor}{rgb}{0.956,0.298,0.235}
\definecolor{citecolor}{HTML}{1976D2}
\hypersetup{
    colorlinks=true,
    linkcolor=citecolor,
    filecolor=magenta,
    urlcolor=cyan,
    citecolor=citecolor,
}

\usepackage{rotating}
\usepackage{enumitem}
\usepackage[capitalize,noabbrev]{cleveref}
\crefname{section}{Sec.}{Secs.}
\Crefname{section}{Section}{Sections}
\Crefname{table}{Table}{Tables}
\crefname{table}{Tab.}{Tabs.}

\definecolor{deemph}{gray}{0.6}

\reportnumber{001}

\renewcommand{\today}{}

\title{\centering Benchmarking the Thinking Mode of Multimodal Large Language Models in Clinical Tasks}
\author[*]{
\small
Jindong Hong$^{1,2}$ \quad Tianjie Chen$^{1}$ \quad
Lingjie Luo$^{1}$ \quad Chuanyang Zheng$^{3}$ \quad  Ting Xu$^{3}$ \quad Haibao Yu$^{4}$ 

\quad Jianing Qiu$^{5}$ \quad
Qianzhong Chen$^{6}$ \quad Suning Huang$^{6}$   \quad Yan Xu$^{7}$ \quad Yong Gui$^{1}$ \quad Yijun He$^{1}$ 

Jiankai Sun$^{1,\dag}$ 
\\

\small
$^1$Bytedance \quad $^2$Peking University \quad $^3$The Chinese University of Hong Kong 

$^4$The University of Hong Kong \quad $^5$Mohamed bin Zayed University of Artificial Intelligence \quad 

$^6$Stanford University \quad $^7$University of Michigan\\
\quad \\
\small
$^\dag$Corresponding Author \\
}

\begin{abstract}
A recent advancement in Multimodal Large Language Models (MLLMs) research is the emergence of "reasoning MLLMs" that offer explicit control over their internal thinking processes (normally referred as the "thinking mode") alongside the standard "non-thinking mode". This capability allows these models to engage in a step-by-step process of internal deliberation before generating a final response. With the rapid transition to and adoption of these "dual-state" MLLMs, this work rigorously evaluated how the enhanced reasoning processes of these MLLMs impact model performance and reliability in clinical tasks. This paper evaluates the active "thinking mode" capabilities of two leading MLLMs, Seed1.5-VL and Gemini-2.5-Flash, for medical applications. We assessed their performance on four visual medical tasks using VQA-RAD and ROCOv2 datasets. Our findings reveal that the improvement from activating the thinking mode remains marginal compared to the standard non-thinking mode for the majority of the tasks. Their performance on complex medical tasks such as open-ended VQA and medical image interpretation remains suboptimal, highlighting the need for domain-specific medical data and more advanced methods for medical knowledge integration. 
\end{abstract}
\begin{document}

\flushbottom
\maketitle

\section{Introduction}
Thinking is widely acknowledged as a cornerstone of human intelligence, enabling structured problem-solving, logical inference, and complex decision making across diverse tasks. In the context of large language models (LLMs)~\citep{thirunavukarasu2023large,zheng2025understanding,zheng2025sas,ren2025decoder,xu2025seqpo,ye2024reasoning,zheng2024dape} and multimodal large language models (MLLMs)~\citep{qiu2024application}, this refers to the model's ability to explicitly articulate its internal processing steps, verify its logic, and adjust its thought process before arriving at a final answer~\citep{sun2023survey}. Recent MLLMs, such as Gemini 2.5~\citep{Kavukcuoglu2025Gemini}, Seed1.5-VL~\citep{guo2025seed1}, provide more flexibility of this ability by direct control of the thinking mode~\citep{zhang2025synapseroute}. The user can choose to open or close the thinking mode based on its demand. In addition, Gemini-2.5-flash can manually set the number of tokens used for thinking. Although longer sequences of thinking processes often lead to improved accuracy, they also result in an increase of inference latency and substantial computational overhead, a phenomenon often referred to as the "overthinking problem" \citep{sui2025stop}. Experiments on MATH500 dataset show that thinking only shows its supremacy on challenging problems \citep{zhang2025adaptthink}, while non-thinking mode responds both quickly and accurately on simple ones. Therefore, it is necessary to investigate the appropriate usage scenarios of the thinking mode, particularly in the medical field where task difficulty varies significantly, ranging from simple tasks like discharge summary~\citep{patel2023chatgpt} to complex scenarios such as AI medical consultant~\citep{fan2024ai}. \par

Although MLLMs with thinking mode evolve rapidly, the effectiveness of their thinking ability in medical tasks has not been fully explored. Firstly, most medical benchmarks focus on the comparison of different models, while the performance difference of the same model with and without the thinking mode remains unexplored. Secondly, only a limited number of benchmarks have compared the differences within the same model; however, these comparisons are restricted to single tasks such as question answering~\citep{nori2024medprompt} and clinical document classification~\citep{mustafa2025can}, the influence of task complexity on the advantages of thinking mode has not been examined. \par
To address the above challenges, we aim to answer two questions: To what extent does the thinking mode enhance the performance of MLLMs in solving visual-language medical tasks, especially for those complex ones? And under which specific medical tasks does thinking mode yield the most substantial performance advantages compared with non-thinking mode? Specifically, we devise four medical tasks (close-ended visual question answering (VQA), open-ended VQA, concept detection, and caption prediction) and conduct tests on two prominent MLLMs with thinking and non-thinking mode control (Seed1.5-VL and Gemini-2.5-Flash). \par
After extensive experiments, our findings can be summarized as: (1) activating the thinking mode generally improves performance, but the improvement gain is below expectation and their advantages only become noticeable when the complexity of medical tasks increases. (2) the consistency of the model output is lower in the thinking mode than in the non-thinking mode. (3) overall performance of leading MLLMs on highly complex medical tasks remains suboptimal, highlighting the need for integrating domain-specific medical data and more advanced multimodal learning methods. \par

\section{Related Works}

\subsection{MLLMs in the Medical Field}
MLLMs are rapidly catalyzing transformative changes across various facets of healthcare by effectively integrating and processing diverse medical data types, including textual patient records, medical images, audio recordings of consultations, and even video data from surgical procedures~\citep{qiu2023large,hong2025diagnosing,qiu2025emerging}. In particular, MLLMs are being developed to generate accurate and comprehensive medical reports from various imaging modalities, such as X-ray~\citep{thawkar2023xraygpt}, CT~\citep{zhang2025mepnet}, MRI~\citep{lei2024autorg} and 3D scan images~\citep{li2025towards}. By pre-training or fine-tuning on the medical data, MLLMs can be used as chatbots to answer visual medical questions or make diagnosis from multimodal data. Foundation models like HuatuoGPT-Vision~\citep{chen2024huatuogpt} and LLaVA-Med~\citep{li2023llava} can perform VQA tasks on various medical images. Fine-tuned MLLMs such as SkinGPT-4~\citep{zhou2023skingpt}, can make diagnosis based on the skin image uploaded by the user in an interactive way. Due to its strong feature extraction ability, MLLMs can also be trained to complete traditional supervised tasks. For example, by pre-training on text and image modalities, Med-MLLM~\citep{liu2023medical} can be used for COVID-19 diagnosis and prognosis tasks.  MLLMs are being proposed as powerful tools to assist or even augment expert work in digital surgery~\citep{lam2024foundation}. Models like SurgicalGPT~\citep{seenivasan2023surgicalgpt}, trained on surgical video data, can answer real-time questions in operating room settings, generate detailed surgical reports, and provide crucial decision support for subsequent procedural steps. MLLMs has also been used to assist dietary assessment ranging from food recognition, volume estimation, to nutrition quantification and recommendation~\citep{lo2024dietary}

\subsection{Reasoning MLLMs and Thinking Mode Control}
Recent reasoning MLLMs, such as OpenAI-o3~\citep{OpenAI2025o3}, Gemini2.5 and Seed1.5-VL, all provide the powerful thinking ability to generate chain of thought before responding. To acquire this ability, these models normally need large-scale reinforcement learning and carefully-designed training recipe.
While longer sequences of thinking process often lead to improved accuracy, they also result in an increase in inference latency and substantial computational overhead, a phenomenon often referred to as the "overthinking problem". Ma et al.~\citep{ma2025reasoning} demonstrates that when controlling the number of tokens or latency, Nothinking (deliberately skipping the thinking process of reasoning models by prompt) is more effective compared with thinking. Zhang et al.~\citep{zhang2025adaptthink} conducted experiments on the MATH500 dataset showing that thinking only shows its supremacy on challenging problems, while non-thinking mode responses both quickly and accurately on simple ones.
Some multimodal foundation models, such as Gemini 2.5, Qwen3 \citep{yang2025qwen3} and Seed1.5-VL, provide the controllability of thinking mode, which means users can dynamically control if using thinking mode. Furthermore, Gemini 2.5 and Qwen3 provide more freedom on how much token can be used for thinking (also known as "thinking budget"). This gives more space for users to dynamically choose the appropriate mode to balance speed, cost and performance.

\section{Experiments}

\subsection{Dataset Preparation}
In our study, we utilize two publicly available datasets to evaluate the performance of dual-state MLLMs: VQA-RAD for visual question answering and ROCOv2 for image captioning and concept detection.

\subsubsection{VQA-RAD}
The VQA-RAD~\citep{lau2018dataset} (Visual Question Answering in Radiology) dataset is a clinically focused collection of medical images paired with question-answer sets, designed to assess models’ abilities to interpret radiological images in response to clinically relevant questions. It comprises 315 radiology images, including modalities such as Computed Tomography (CT), Magnetic Resonance Imaging (MRI), and X-ray, evenly distributed across three anatomical regions: head, chest, and abdomen. These images are associated with 3,515 question-answer (QA) pairs, averaging approximately 10 QA pairs per image. The questions are categorized into 11 distinct types, reflecting clinically relevant tasks: abnormality, attribute, modality, organ system, color, counting, object/condition presence, size, plane, positional reasoning, and a general “other” category. This categorization provides insight into the types of queries clinicians commonly pose. \par
In this work, we leverage the entire VQA-RAD dataset to evaluate MLLM performance on two distinct visual question answering tasks: \par
\begin{itemize}
    \item Closed-ended VQA: This task involves questions with a limited set of possible answers, specifically “yes/no” questions in VQA-RAD. The model’s objective is to select the correct answer based on the visual information in the corresponding medical image.
    \item Open-ended VQA: This task requires the model to generate concise free-text answers to questions about a given medical image, demanding deeper image understanding and natural language generation capabilities, as answers are not restricted to predefined options.
\end{itemize}

For closed-ended VQA, we primarily report accuracy, defined as the proportion of correctly predicted answers relative to the total number of questions. To ensure stable results, we conduct three runs for each closed-form VQA task and report consistency as a reference metric, measuring the proportion of identical outputs across the three runs.
For open-ended VQA, where answers are free-form text, we report traditional NLP metrics such as BLEU and ROUGE-L. Given the variability in model outputs, semantic correctness is prioritized. We employ an LLM judge to evaluate whether the model’s output matches the ground-truth answer, returning a True or False judgment for each question. Consequently, the score of accuracy is also reported for open-form VQA.

\subsubsection{ROCOv2}

The ROCOv2~\citep{ruckert2024rocov2} (Radiology Objects in Context version 2) dataset is a comprehensive multimodal dataset for medical image analysis, representing an enhanced and expanded version of the original ROCO dataset. It contains 79,789 radiological images across various modalities, including CT scans, X-rays, ultrasounds, and MRIs, paired with associated medical concepts (primarily UMLS CUIs) and their original textual captions. In this study, we use the ROCOv2 dataset to assess MLLM performance in generating medical image captions. Due to rate limitations, we evaluate the first 2,000 images from the test set, which maintain a similar distribution to the full test set.
We designed two tasks to reflect varying levels of complexity:
\begin{itemize}
\item Concept Detection: This task requires the model to predict relevant medical concepts, such as UMLS CUIs, imaging modalities, or anatomical regions, directly from the visual information in the radiological images. It focuses on identifying key concepts without requiring semantic connections between them.
\item Caption Prediction: This task demands that the model fully interprets the image and captures the semantic relationships among multiple medical concepts to generate a coherent caption.
\end{itemize}
For concept detection, we employ another LLM to count the number of CUIs recalled in the model’s output. We report two metrics: 1) Simple Accuracy: The proportion of model outputs with the correct number of recalled CUIs; 2) Total Concept Recall: The ratio of CUIs recalled by the model to the total CUIs in the ground truth.

For caption prediction, we report traditional NLP metrics such as ROUGE-L and BLEU. Additionally, we use an LLM judge to assess caption quality, rating model outputs on a scale from 0 to 3 (0 indicates no semantic correlation with the reference caption, and 3 indicates high semantic correlation or equivalent meaning). The reliability of the LLM judge is validated through human evaluation by three medical imaging professionals, with 70\% of the LLM judge’s scores aligning with human assessments.

\begin{table}[htbp]
\centering
\caption{Overview of VQA-RAD and ROCOv2 datasets}
\resizebox{\linewidth}{!}{
\begin{tabular}{p{2cm}p{2.5cm}p{3.5cm}p{3cm}p{3.5cm}}
\toprule
\textbf{Dataset} & \textbf{Size} & \textbf{Image Modality} & \textbf{Tasks} & \textbf{Evaluation Metric} \\
\midrule
VQA-RAD & 
\parbox{2.5cm}{
\begin{itemize}[leftmargin=*,noitemsep,topsep=0pt]
\item 1,192 close-form VQA
\item 1,053 open-form VQA
\end{itemize}
} & 
\parbox{3.5cm}{
\begin{itemize}[leftmargin=*,noitemsep,topsep=0pt]
\item 104 head axial single-slice CTs or MRIs
\item 107 chest X-rays
\item 104 abdominal axial CTs
\end{itemize}
} & 
\parbox{3cm}{
\begin{itemize}[leftmargin=*,noitemsep,topsep=0pt]
\item Close-ended VQA: yes/no questions
\item Open-ended VQA: free-text answers
\end{itemize}
} & 
\parbox{3.5cm}{
\begin{itemize}[leftmargin=*,noitemsep,topsep=0pt]
\item Close-ended: Accuracy
\item Open-ended: LLM judge accuracy
\end{itemize}
} \\
\midrule
ROCOv2 & 2,000 images & 
\parbox{3.5cm}{
\begin{itemize}[leftmargin=*,noitemsep,topsep=0pt]
\item 746 CTs
\item 515 X-rays
\item 326 ultrasounds
\item 306 MRIs
\item 82 angiograms
\item 25 other modalities (mainly PET and PET-CT)
\end{itemize}
} & 
\parbox{3cm}{
\begin{itemize}[leftmargin=*,noitemsep,topsep=0pt]
\item Concept detection
\item Caption prediction
\end{itemize}
} & 
\parbox{3.5cm}{
\begin{itemize}[leftmargin=*,noitemsep,topsep=0pt]
\item Concept detection: Accuracy, total concept recall
\item Caption prediction: LLM judge score
\end{itemize}
} \\
\bottomrule
\end{tabular}
}
\label{tab:datasets_overview1}
\end{table}


\subsection{Experimental Setup}

\subsubsection{MLLM Setup}
We conducted our experiments using two state-of-the-art MLLMs with direct control over thinking capabilities: Seed1.5-VL and Gemini-2.5-Flash. Other models with thinking ability control were excluded from this study. Specifically, the Qwen3 series does not currently support image inputs, and Gemini-2.5-Pro lacks the option to disable its thinking ability. At the time of our experiments, the latest versions used were Doubao-1.5-Thinking-Vision-Pro-0428 for Seed1.5-VL and Gemini-2.5-Flash-Preview-04-17 for Gemini-2.5-Flash. We deploy our experiments via API provided by Volcano Engine and Azure. To optimize reasoning performance, we imposed no restrictions on the number of reasoning tokens used in thinking mode.

\subsubsection{System Prompt}
Our system prompt setting for each task is explicit and provide few example outputs to control output style.

\begin{itemize}
    \item Close-ended VQA

\begin{tcolorbox}
You are a medical doctor and expert in medical imaging. 

Please read the uploaded image and answer the question.

Note:

1. We expect the answer to only use "yes" or "no", no other words

2. Respond only in English.

Question:
\end{tcolorbox}
    \item Open-ended VQA
\begin{tcolorbox}
You are a medical doctor and expert in medical imaging. \\
Please read the uploaded image and answer the question. \\
Note:  \\
1. We expect the response to be short and concise \\
2. Respone only in English. \\
3. Be careful about the potential flips on the image, especially left-right flips.\\
\\
Example 1:\\
Question: what type of imaging is this\\
Answer: CT\\
\\
Example 2:\\
Question: where are the brain lesions located\\
Answer: left hemisphere\\
\\
Example 3:\\
Question: what is the primary abnormality in this image\\
Answer: ring enhancing lesion in the right occipital lobe\\
\\
Question: 
\end{tcolorbox}
    \item Caption Prediction \& Concept Detection
    \begin{tcolorbox}
You are a medical doctor and expert in medical imaging. \\
\\
Please read the uploaded image and generate the corresponding caption.\\
\\
Note: \\
1. We expect the responsed to be short and concice.\\
2. Your answer should be precise and free of incomplete or incorrect biomedical or clinical information.\\
\\
Example Captions:\\
1. Anteroposterior pelvic radiograph of a 30-year-old female diagnosed with Ehlers- Danlos Syndrome demonstrating fusion of pubic symphysis and both sacroiliac joints (anterior plating, bone grafting and sacroiliac screw insertion)\\
2. CT scan image for lung cancer.\\
3. A giant retroperitoneal tumor.\\
4. Early axial T2-weighted MRI.\\
5. Neck and head computed tomography image showing left odontogenic infection.
\end{tcolorbox}
\end{itemize}

\subsection{Main Results}
\begin{figure}[ht!]
\centering

\begin{subfigure}[b]{0.9\linewidth}
    \centering
    \includegraphics[width=\linewidth]{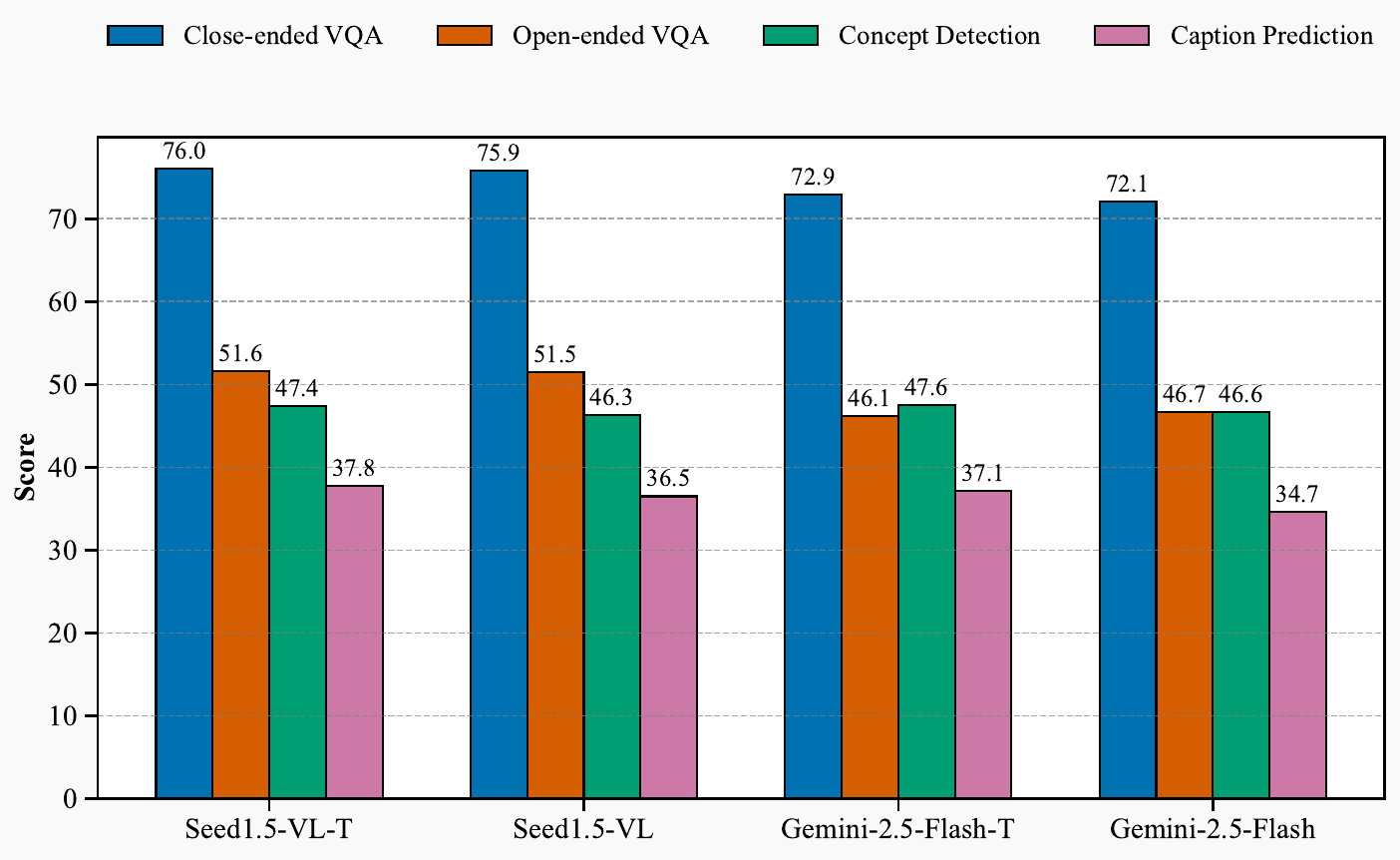}
    \caption{Model performance across different tasks (T: thinking mode).}
    \label{fig:model_performance}
\end{subfigure}

\vspace{0.6em}

\begin{subfigure}[b]{0.9\linewidth}
    \centering
    \includegraphics[width=\linewidth]{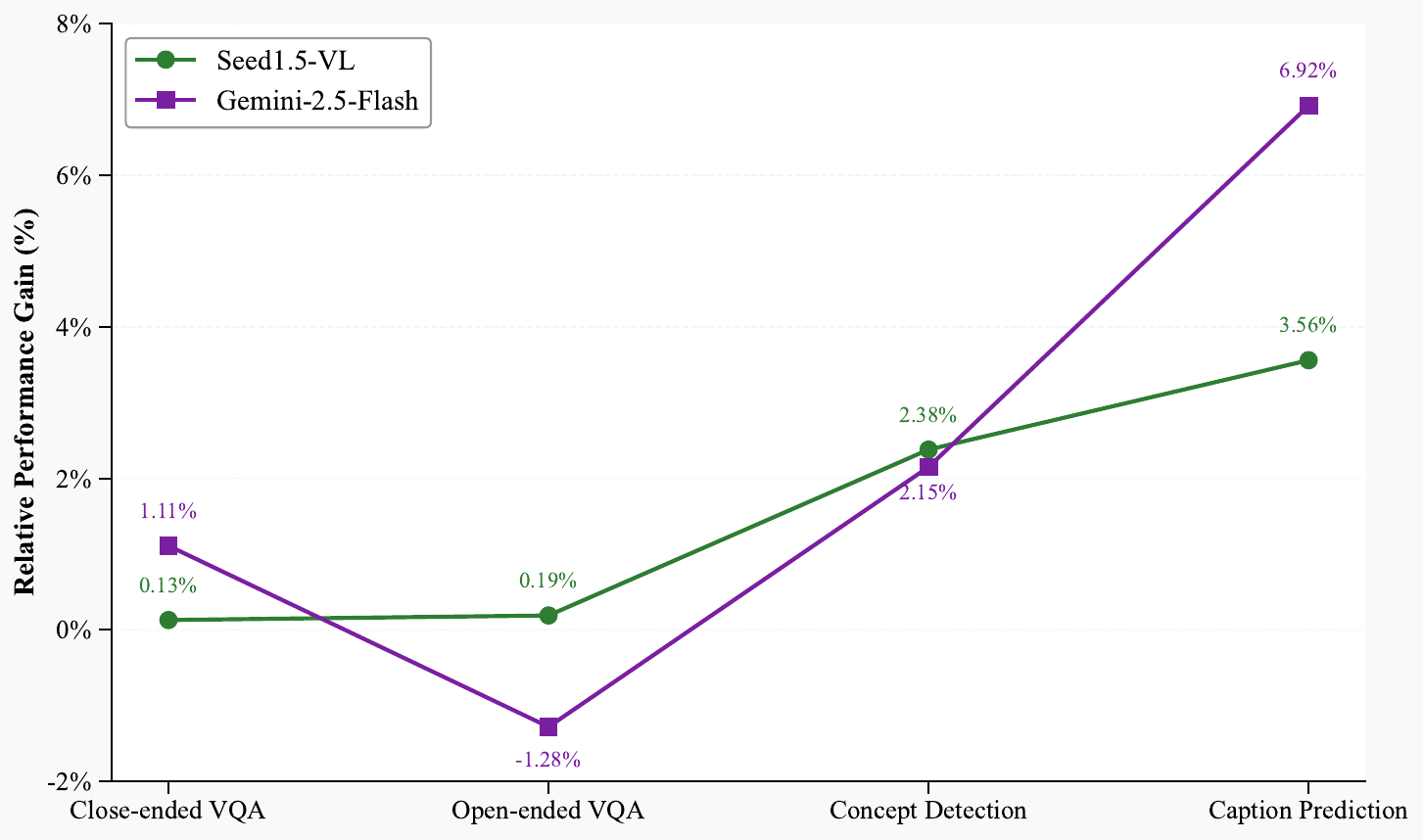}
    \caption{Thinking mode performance gain.}
    \label{fig:thinking_mode_performance_gain}
\end{subfigure}

\caption{(a) Model performance across tasks; (b) Thinking mode relative performance gain.}
\label{fig:combined_model_performance}
\end{figure}

Figure~\ref{fig:model_performance} shows the performance comparison of different models across visual medical tasks. Figure~\ref{fig:thinking_mode_performance_gain} shows the relative increase in thinking mode performance compared to non-thinking mode. In summary: \par
\textbf{Model performance decreases progressively as task complexity increases.} Both models show a significant decreasing trend in model performance as task complexity increases, with scores decreasing progressively from Close-ended VQA to Open-ended VQA, Concept Detection, and Caption Prediction. \par
\textbf{The advantage of thinking over non-thinking only becomes noticeable when the task is more complex} For two VQA tasks, thinking mode shows limited advantages, even a 1.28\% relative decline for Gemini-2.5-Flash on open-ended VQA. For two image caption tasks, thinking mode shows slightly clear advantages. \par
\textbf{Performance of both models on complex medical tasks remains suboptimal.} Except for the Close-ended VQA, the average score of other tasks is close or lower than 50\%, suggesting a potential need for domain-specific training or fine-tuning to enhance accuracy. Although the thinking ability improves model performance, the incremental gains are insufficient to meet usability standards for highly complex tasks.

\subsection{Fine-grained Analysis of Thinking Mode}

\subsubsection{Does VQA performance vary among different question types?}

As Table~\ref{tab:datasets_overview} shows, for the VQA-RAD dataset, all questions fall into 11 categories defined by clinicians,  reflecting the corresponding tasks naturally occurring about radiology images. \newpage

\begin{table}[ht]
\centering
\caption{Descriptive information of VQA-RAD dataset}
\begin{tabular}{p{2.5cm}p{2cm}p{3.5cm}p{3cm}p{1.cm}p{1.5cm}}
\toprule
\textbf{Question Type} & \textbf{Abbreviation} & \textbf{Description} & \textbf{Example Question} & \textbf{Count} & \textbf{Proportion} \\
\midrule		
object/condition presence & pres & if certain object/condition present & is there evidence of an aortic aneurysm & 813 & 36.21\% \\
\midrule
positional reasoning & pos & position or location of an object or organ &	where is the abnormality & 322 & 14.34\% \\
\midrule
abnormality	& abnormality & if abnormality shows in the image & is the lung normal & 204 & 9.09\% \\
\midrule
other & other & other questions out of listed categories & how is the patient oriented & 196 & 8.73\%  \\
\midrule
modality & modality & modality of image & is this a MRI of the chest & 185 & 8.24\% \\
\midrule
size & size & size of an object & is the spleen normal size
& 175 & 7.80\% \\
\midrule
plane & plane & orientation of image slicing through the body & is this an anterior or posterior image & 120 & 5.35\% \\
\midrule
attribute other & attrib & other types of description questions & are the hilar soft tissue densities symmetric & 93 & 4.14\% \\
\midrule
organ system & organ & category of a system & what organ system is pictured & 59 & 2.63\% \\
\midrule
color & color & signal intensity & does the csf have high signal intensity & 54 & 2.41\% \\
\midrule
counting & count & quantity of objects & how many lesions are in the spleen & 24 & 1.07\% \\
\bottomrule
\end{tabular}
\label{tab:datasets_overview}
\end{table}

\begin{figure}[ht!]
\centering

\begin{subfigure}[b]{0.8\linewidth}
    \centering
    \includegraphics[width=\linewidth]{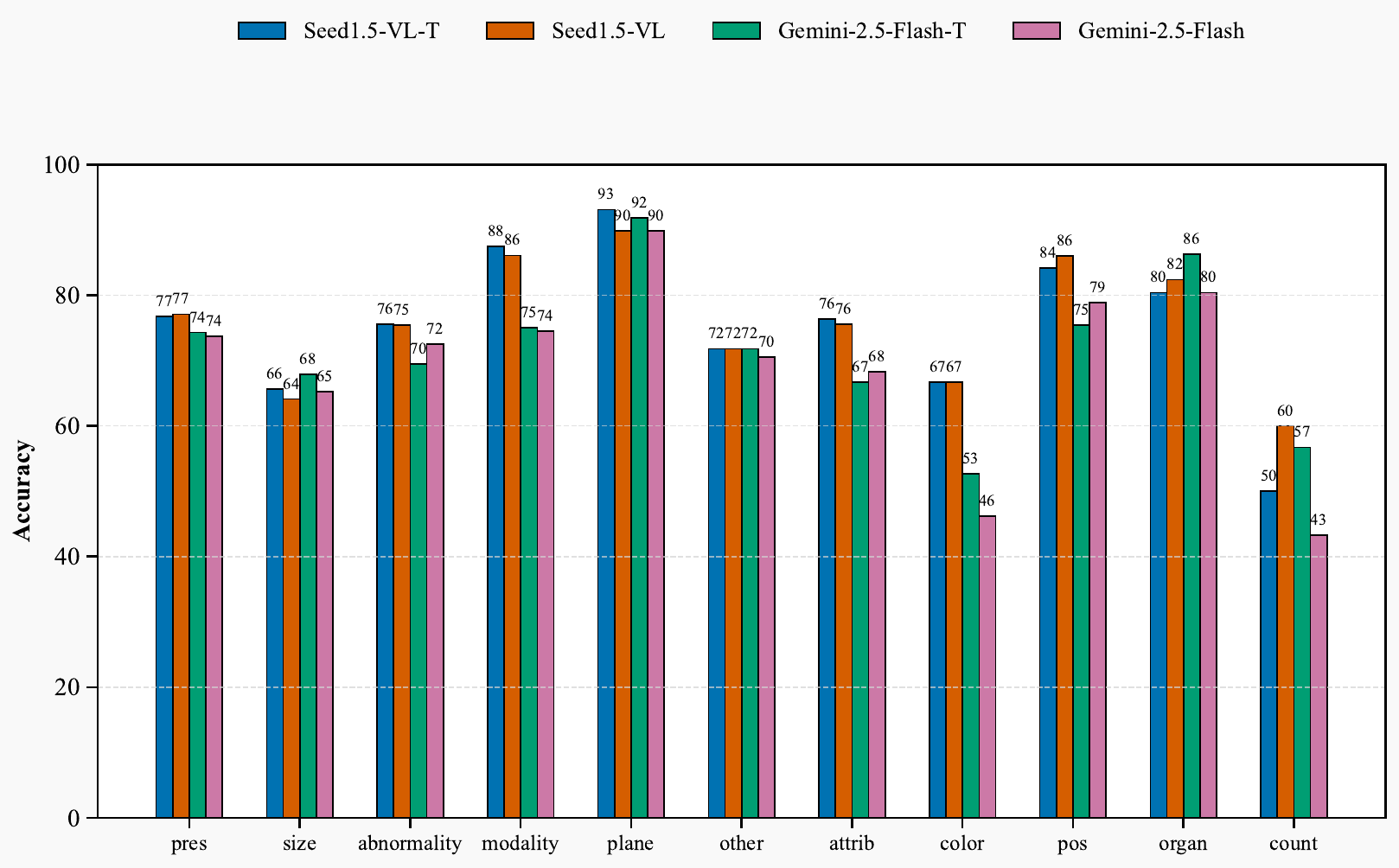}
    \caption{Close-ended VQA: Model performance across question types}
    \label{fig:close_vqa_model_performance}
\end{subfigure}

\vspace{0.5em}

\begin{subfigure}[b]{0.8\linewidth}
    \centering
    \includegraphics[width=\linewidth]{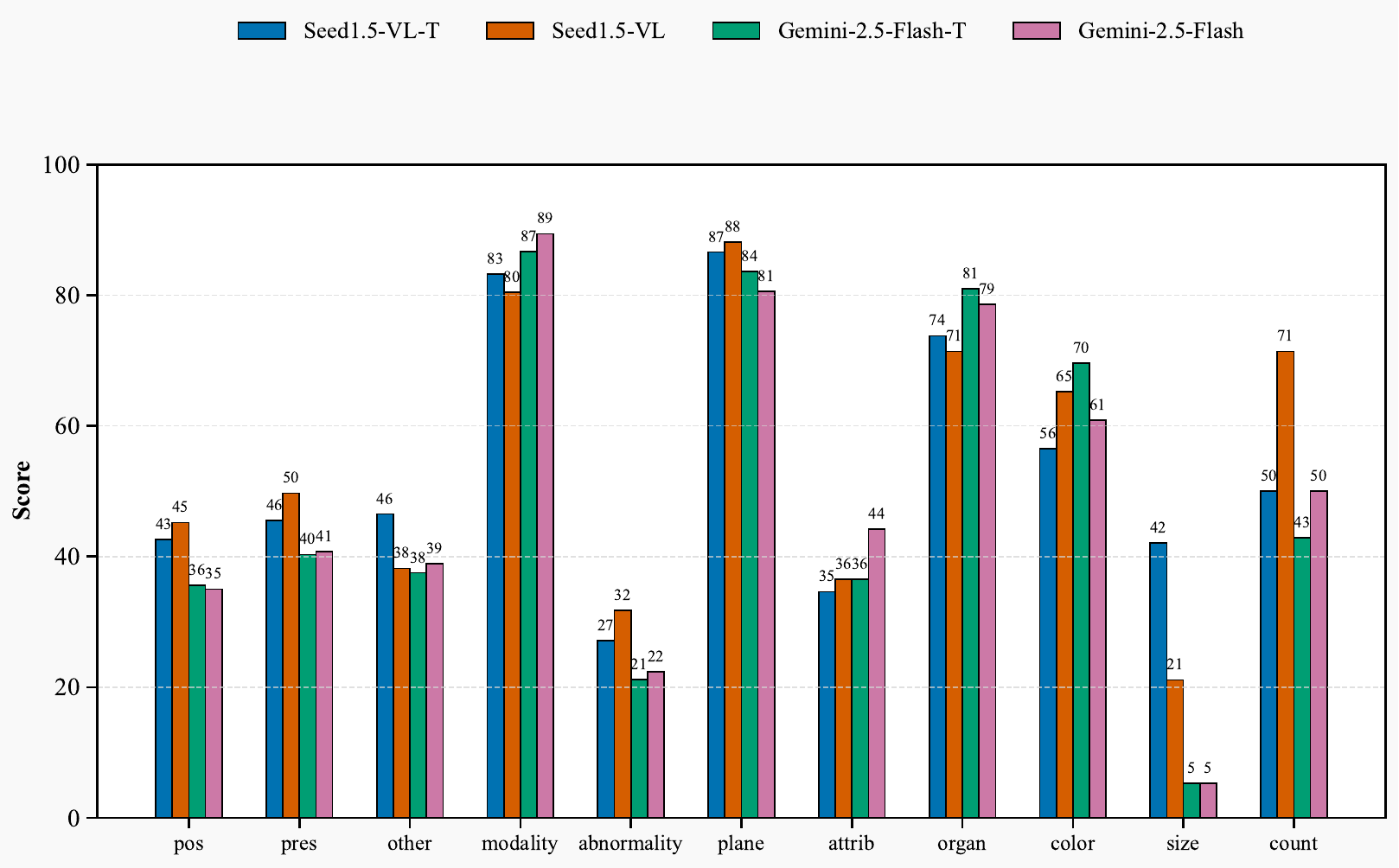}
    \caption{Open-ended VQA: Model performance across question types}
    \label{fig:open_vqa_model_performance}
\end{subfigure}

\caption{(a) Close-ended VQA; (b) Open-ended VQA performance across question types.}
\label{fig:vqa_model_performance_combined}
\end{figure}

For close-ended VQA, the main question type are object/condition presence, accounting for 52.18\% of questions as shown in Table \ref{tab:close_ended_vqa}. For all 11 question types, the thinking mode of Seed1.5-VL leads 5 times, and that of Gemini-2.5-Flash leads 8 times as shown in Figure \ref{fig:close_vqa_model_performance} . Notably, all model achieve high accuracy on "plane" and "modality", which are also the question types thinking mode exceed non-thinking mode.

\begin{table}
\centering
\caption{Close-ended VQA Question Type Distribution}
\begin{tabular}{ccc}
\toprule
\textbf{Type} & \textbf{Count} & \textbf{Proportion}  \\
\midrule			
pres & 622 & 52.18\% \\
size & 156 & 13.09\% \\
abnormality & 119 & 9.98\% \\
modality & 72 & 6.04\% \\
plane & 53 & 4.45\% \\
other & 52 & 4.36\% \\
attrib & 41 & 3.44\% \\
color & 31 & 2.60\% \\
pos & 19 & 1.59\% \\
organ & 17 & 1.43\% \\
count & 10 & 0.84\% \\
\bottomrule
\end{tabular}
\label{tab:close_ended_vqa}
\end{table}

For open-ended VQA, the main question types are position, object/condition presence, accounting for 28.77\% and 18.14\% of questions as shown in Table \ref{tab:open_ended_vqa}. For all 11 question types, both the thinking mode of Seed1.5-VL and Gemini-2.5-Flash leads 4 times as shown in Figure \ref{fig:open_vqa_model_performance}. Notably, except "organ", thinking mode shows no consistent supremacy on most categories. 

\begin{table}
\centering
\caption{Open-ended VQA Question Type Distribution}
\begin{tabular}{ccc}
\toprule
\textbf{Type} & \textbf{Count} & \textbf{Proportion}  \\
\midrule
pos	& 303 & 28.77\%  \\
pres & 191 & 18.14\% \\
other & 144 & 13.68\% \\
modality & 113 & 10.73\% \\
abnormality & 85 & 8.07\% \\
plane & 67 & 6.36\% \\
attrib & 52 & 4.94\% \\
organ & 42 & 3.99\% \\
color & 23 & 2.18\% \\
size & 19 & 1.80\% \\
count & 14 & 1.33\% \\
\bottomrule
\end{tabular}
\label{tab:open_ended_vqa}
\end{table}

\begin{table}[h]
\centering\tabcolsep 1.9cm
\caption{Distribution of modalities for ROCOv2}
\resizebox{\linewidth}{!}{
\begin{tabular}{ccc}
\toprule
\textbf{Modality} & \textbf{Count} & \textbf{Proportion}  \\
\midrule
CT & 746 & 37.30\% \\
X-ray & 515 & 25.75\% \\
Ultrasound & 326 & 16.30\% \\
MRI & 306 & 15.30\% \\
angiogram & 82 & 4.10\% \\
others & 25 & 1.25\% \\
\bottomrule
\end{tabular}
}
\label{tab:distr_rocov2}
\end{table}

\subsubsection{Does the model performance on ROCOv2 vary across different image modalities?}

Table~\ref{tab:distr_rocov2} shows the distribution of modalities for our ROCOv2 evaluation dataset. CT, X-Ray, ultrasound and MRI are 4 main modalities, accounting for 94.65\% of images. The remaining are rare modalities like angiogram and pet-ct.

\begin{figure}[!ht]
\centering

\begin{subfigure}[b]{0.8\linewidth}
    \centering
    \includegraphics[width=\linewidth]{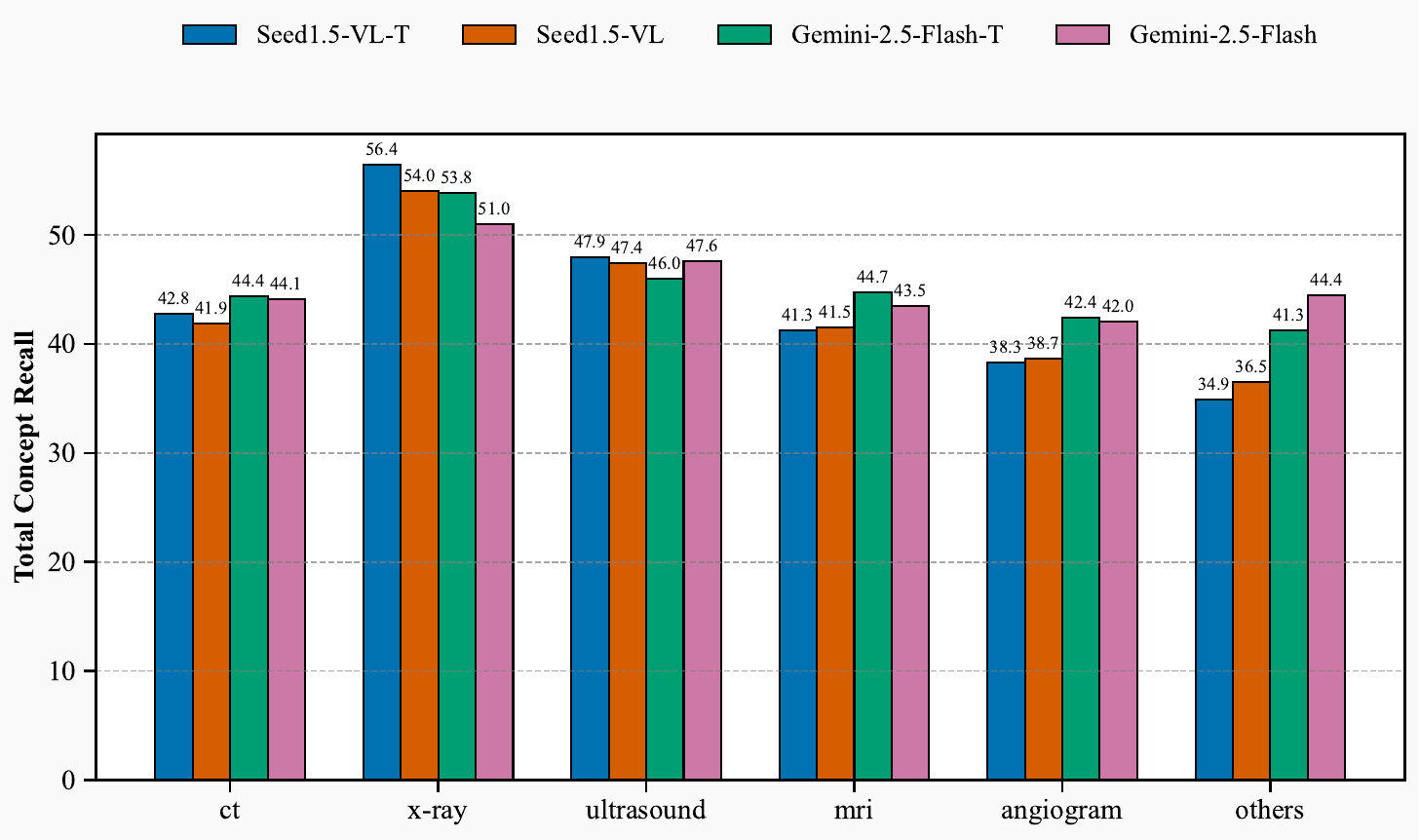}
    \caption{Concept Detection Across Modalities}
    \label{fig:concept_detection}
\end{subfigure}

\vspace{0.5em}

\begin{subfigure}[b]{0.8\linewidth}
    \centering
    \includegraphics[width=\linewidth]{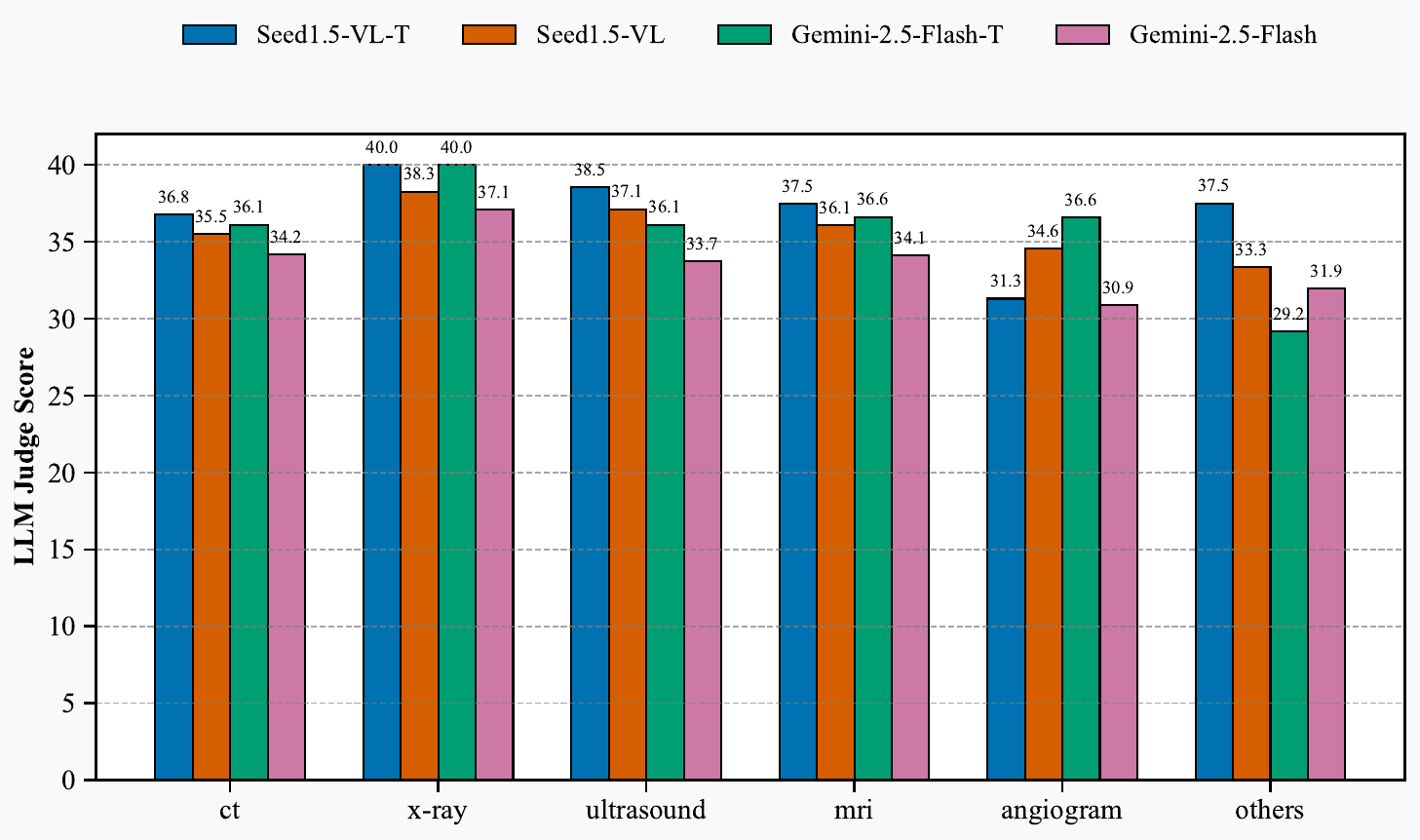}
    \caption{Caption Prediction Across Modalities}
    \label{fig:caption_prediction}
\end{subfigure}

\caption{(a) Concept detection; (b) Caption prediction across modalities.}
\label{fig:combined_concept_caption}
\end{figure}

 The models demonstrated varying performance across imaging modalities, with the highest average scores achieved for X-ray, followed by ultrasound, while MRI and CT showed comparable performance (X-ray > ultrasound > MRI $\approx$ CT) as shown in Figure \ref{fig:concept_detection}. Relatively lower performance was observed for angiography and other rare modalities (e.g., PET-CT), likely reflecting the limited representation of these imaging modalties in the training dataset, which may have resulted in insufficient learning of their distinctive features. \par
The performance gain of the thinking mode over non-thinking mode also seems to be modality-dependent. As shown in Figure \ref{fig:caption_prediction}, for the caption prediction task, the thinking mode shows advantages over the non-thinking mode in only CT, X-ray, Ultrasound, and MRI modalities.

\newpage

\subsubsection{How thinking mode affects model output consistency?}

\begin{figure}
     \centering
    \includegraphics[width=0.9\linewidth]{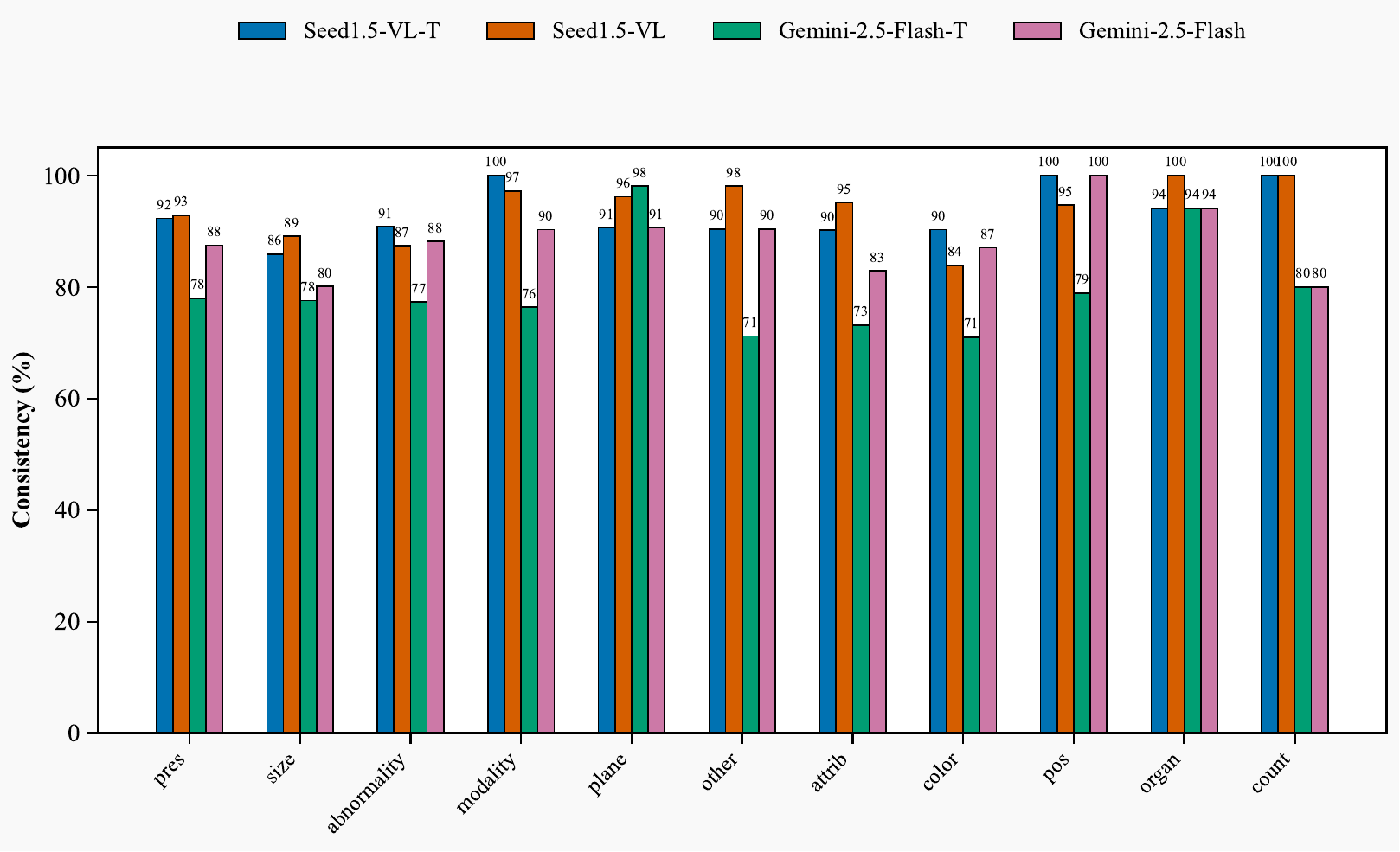}
    \captionof{figure}{Consistency across Problem Categories}
    \label{fig:model_consistency}
\end{figure}

Beyond performance metrics, model output consistency is another important metric, which measures if LLMs can produce the same or similar outputs when supplied with inputs that are semantically equivalent. In the medical domain, high consistency ensures the reliability of model and minimize the risk of incorrect diagnoses or treatment recommendations. We measure the consistency by checking whether MLLMs output the same results on 3 rounds of close-ended VQA tasks. To ensure fair comparison, the default model output temperature is set same as 0.8 for each model. Our findings, summarized in Table~\ref{tab:overall_consistency}, indicate that for both Gemini-2.5-Flash and Seed1.5-VL, \textbf{thinking mode resulted in lower model output consistency compared to the non-thinking mode. This suggests that the thinking processes may introduce more randomness.} Specifically, Gemini-2.5-Flash exhibited a more substantial decrease in consistency, with an 8.8\% drop when the thinking mode was activated. In contrast, Seed1.5-VL demonstrated greater robustness in this regard, showing only a 0.84\% drop in consistency. Figure~\ref{fig:model_consistency} shows the consistency across different problem categories, Gemini-2.5-Flash shows decrease of consistency in almost each category except "plane", whereas Seed1.5-VL demonstrates random advantages of thinking and non-thinking modes over each other.

\begin{table}[]
\centering
\caption{Overall Consistency on Close-ended VQA}
\label{tab:overall_consistency}
\begin{tabular}{ccc}
\toprule
\textbf{Model} & \textbf{Consistency } \\
\midrule
Seed1.5-VL-T & 91.69\% \\
Seed1.5-VL & 92.53\% \\
Gemini-2.5-Flash-T  & 78.27\% \\
Gemini-2.5-Flash & 87.08\% \\
\bottomrule
\end{tabular}
\end{table}

\subsubsection{How does thinking tokens change as task complexity varies?}

Considering the average thinking tokens for each task, we can see a clear trend that model tends to think more when task complexity increases. From Close-ended VQA to Open-ended VQA, both Seed1.5-VL-T and Gemini-2.5-Flash-T make more effort on thinking. While in caption generation task, Seed1.5-VL and Gemini-2.5-flash show totally different trends, Seed1.5-VL-T treated it as a simple task and spent few tokens on thinking, while Gemini-2.5-Flash-T continues to expand thinking effort as shown in Table \ref{tab:average_thinking_tokens}. We further conducted an experiment by modifying the prompt templates and forced Seed1.5-VL-T to "think carefully before you generate caption, do not ignore details of images". The average thinking tokens remain 141.6, which may \textbf{reflect the difference between models' innate understanding of task complexity}.

\begin{table}[htbp]
\centering
\caption{Average thinking tokens across tasks}
\begin{tabular}{p{3.5cm}p{2cm}p{3.5cm}p{3.5cm}}
\toprule
\textbf{Model} & \textbf{Close-ended VQA} & \textbf{Open-ended VQA} & \textbf{Concept Detection} \\
\midrule
Seed1.5-VL-T & 75.2 & 273.1 & 105.5 \\
Gemini-2.5-Flash-T & 381.4 & 756.2 & 1167.9 \\
\bottomrule
\end{tabular}
\label{tab:average_thinking_tokens}
\end{table}

\subsubsection{Model Speed VS. Performance}

Besides performance improvement, speed is another factor that affects usability. We define the latency of LLM output as: from when a prompt is originally entered by the user to when they receive the completed output from the model, measured by second. Table~\ref{tab:model_latency_and_performance} shows the latency and performance score for all tasks. For open-ended VQA, the thinking mode has a long delay and brings very little improvement or even a decline in performance. 

\begin{table}[htbp]
\centering
\caption{Comparison of model latency and performance}
\resizebox{\linewidth}{!}{
\begin{tabular}{p{3.3cm}p{1cm}p{1cm}p{1cm}p{1cm}p{1cm}p{1cm}p{1cm}p{1cm}}
\toprule
& \multicolumn{2}{c}{\textbf{Close-ended VQA}}	& 	\multicolumn{2}{c}{\textbf{Open-ended VQA}}	& \multicolumn{2}{c}{\textbf{Concept Recall}} & 	\multicolumn{2}{c}{\textbf{Caption Prediction}} \\
\midrule
Model	& Latency& 	Score&	Latency	&Score	&Latency	&Score&	Latency&	Score \\
\hline
Seed1.5-VL-T&	2.59&	76.03&	10.48&	51.57&	4.09&	47.42&	4.09&	37.78\\
Seed1.5-VL&	0.94&	75.87&	0.80	&51.47	&0.95&	46.32&	0.95&	36.50\\
Gemini-2.5-Flash-T	&4.06&	72.93&	6.62&	46.15	&9.60&	47.57&	9.60&	37.12\\
Gemini-2.5-Flash&	1.56&	72.12&	1.55&	46.72&	1.75&	46.63&	1.75	&34.68\\
\bottomrule
\end{tabular}
}
\label{tab:model_latency_and_performance}
\end{table}

\section{Discussion}

This work has several limitations. The first is \textbf{limited dataset coverage}. The evaluation is based solely on radiology image datasets, which restricts the generalization of our findings. While radiology is a significant domain in medical imaging, it does not encompass the full spectrum of medical image modalities (e.g., pathology slides, dermatology images). As such, our conclusions may not be fully applicable to other medical imaging fields. The second is \textbf{potential data leakage}. Due to the absence of transparent training data documentation for Seed1.5-VL and Gemini-2.5-Flash, it is difficult to ascertain whether the evaluation datasets were part of the models’ pretraining data. This lack of clarity introduces the risk of data leakage, which could inadvertently inflate the performance results of some models. The third is \textbf{reliance on LLM-based evaluation.} While we validated the reliability of LLM-based judge against a subset of expert-annotated ground truths, full-scale human evaluation remains limited. Human expert judgment is still essential, especially for nuanced medical tasks, and future work should incorporate broader clinical assessments to corroborate automated evaluation metrics. The limited accuracy across multiple tasks of the tested "dual-state" MLLMs may stem from their internal capabilities, e.g., lacking proper medical knowledge which can be strengthened through fine-tuning or through agentic AI approaches that augment models with external knowledge~\cite{qiu2024llm}. More importantly, this work suggests that medical reasoning may not be a fix for all tasks. It is critical to understand the complexity of the task before implementing MLLMs as the solution.

\section{Conclusion}
In this paper, we evaluated the active "thinking mode" capabilities of two leading dual-state MLLMs, Seed1.5-VL and Gemini-2.5-Flash, for medical applications. Our experiments show that while thinking mode is a critical factor in enhancing MLLM performance, its advantages over non-thinking mode on various medical tasks are below the expectation. A significant gap persists: current general MLLMs, even with activated thinking modes, continue to perform inadequately in highly complex visual medical scenarios such as medical image interpretation. This indicates that while thinking mode offers an improvement, it is not a complete solution for achieving clinical usability with current general MLLMs.

\section{Supplementary Material}
\subsection{Details of Data Preprocessing}
\paragraph{VQA-RAD}
For VQA-RAD dataset, we use the whole dataset for evaluation. The original dataset has a total of 2248 questions. After dropping 3 duplicate questions, there remain 2245 questions and 314 corresponding images. For question with answer "yes" and "no", we define it as close-form vqa. For remaining ones with no fixed answer, we define it as open-form vqa.

For very few questions that have multiple categories, we classify each of them into the most appropriate category for later analysis
\begin{itemize}
    \item 'pos, pres':  which side is more clearly visualized  --> 'pos'
    \item 'pos, abnormality': 'which lobes demonstrate pathology' --> 'pos'
    \item 'size, color': what is the size and density of the lesion --> 'size'
    \item 'pres, pos': what organ is enlarged --> 'pres'
    \item 'abnormality, pos': what pathology is seen in this image what side --> 'abnormality'
\end{itemize}

\paragraph{ROCOv2}
For ROCOv2, the original images are in png format, which may exceed the maximum file size of model input, we convert them into jpeg to meet the requirement

\subsection{Details of Experiment Setup}
\subsubsection{LLM Setup}
For both Seed1.5-VL and Gemini-2.5-flash, we can explicitly control their thinking ability by changing certain parameters. In API of Gemini-2.5-flash, The "thinkngBudget" parameter can control the behavior of thinking, setting the thinking budget to 0 will disable thinking. In API of Seed1.5-VL, the "thinking.type" parameter can control if thinking ability is enabled, when setting to "disabled", the model will disable thinking ability. Since all tasks require relative few tokens for output, to balance thinking ability and speed, we set maximum output tokens to be 4000 for all tasks, which is adequate for most of scenarios. In very few times, thinking model may fail with this restriction, then we re-run the model until it can generate full output.

We limit the maximum output token to be 4000 in all scenarios.

\subsubsection{LLM Judge Setup}
\newpage
1. Prompt for open-form question judge. 

\begin{tcolorbox}[title=Prompt for open-form question judge]
You are an expert in Natural Language Understanding, specializing in comparing the semantic meaning of texts, particularly in question-answering contexts. \\
\\
Your task is to determine if the `Model Output` has substantially the same semantic meaning as the `Answer`, given the `Question`. \\
\\
**Evaluation Criteria:** \\
\\
Consider the following when making your judgment: \\
1.  **Core Information:** Does the `Model Output` convey the same essential information as the `Answer` in relation to the `Question`? \\
2.  **Accuracy:** Is the `Model Output` factually correct with respect to the `Answer`? \\
3.  **Completeness:** \\
    *   If `Model Output` is a generalization of `Answer` (e.g., `Answer`: "chest x-ray", `Model Output`: "X-ray"), it can be considered to have the same meaning if the generalization still correctly and adequately answers the `Question`. \\
    *   If `Model Output` is more specific but still accurately represents the `Answer`'s core concept (e.g., `Answer`: "cystic lesions", `Model Output`: "multiple cysts"), it can have the same meaning.\\
    *   However, if `Model Output` omits critical details from the `Answer` that are necessary for a complete and accurate response to the `Question`, it does not have the same meaning.\\
4.  **Synonyms and Paraphrasing:** Equivalent terms or rephrased statements that retain the original meaning are considered the same. \\
5.  **Contradiction/Different Concepts:** If the `Model Output` states something different or contradictory to the `Answer`, it does not have the same meaning. \\
\\
**Output Format:** \\
\\
Return your evaluation as a boolean (`True` if they have the same semantic meaning, `False` otherwise) and a brief `Reason`.\\
\\
Format your response exactly as follows:\\
Evaluation: [True/False]\\
Reason: [Your brief explanation]\\
\\
**Examples:**\\
\\
**Example 1:**\\
`Question:` what type of image is this\\
`Answer:` chest x-ray\\
`Model Output:` X-ray\\
Evaluation: True\\
Reason: "X-ray" is a correct and acceptable generalization of "chest x-ray" in the context of identifying the image type.\\
\\
\end{tcolorbox}
\begin{tcolorbox}[title=Prompt for open-form question judge (Cont')]
**Example 2:**\\
`Question:` which organ has the abnormality\\
`Answer:` pancreas\\
`Model Output:` stomach\\
Evaluation: False\\
Reason: "Stomach" is a different organ than "pancreas," making the model output factually incorrect in relation to the answer.\\
\\
**Example 3:**\\
`Question:` what is the brightness in the abdominal aorta\\
`Answer:` atherosclerotic calcification\\
`Model Output:` Calcification\\
Evaluation: True\\
Reason: "Calcification" captures the core meaning of "atherosclerotic calcification" in response to the question about brightness and is an acceptable generalization.\\
\\
**Example 4:**\\
`Question:` describe the lesions in the right kidney\\
`Answer:` cystic lesions\\
`Model Output:` multiple cysts\\
Evaluation: True\\
Reason: "Multiple cysts" is a semantically equivalent description to "cystic lesions" in this medical context; both refer to the same type of pathological finding.\\
\\
---\\
\\
**Now, please evaluate the following:**\\
\\
`Question:` {question}\\
`Answer:` {answer}\\
`Model Output:` {model\_output}\\
\\
Evaluation:\\
Reason:\\
\end{tcolorbox}

2. Prompt for concept recall task judge 

\begin{tcolorbox}[title=Prompt for Concept Matching Task]
You are an expert medical text analyst. Your task is to determine how many concepts from a provided list are substantially represented in a given output string.\\
\\
\textbf{Instructions:}\\
1. You will be given an \texttt{output\_string} and a \texttt{concepts\_list}.\\
2. For each \texttt{concept} in the \texttt{concepts\_list}, carefully evaluate if its core meaning is present in the \texttt{output\_string}.\\
3. When evaluating, consider the following:\\
    \quad \textbullet{} \textbf{Direct Match:} The concept is explicitly stated.\\
    \quad \textbullet{} \textbf{Synonyms \& Abbreviations:} Common medical synonyms or abbreviations (e.g., ``CT'' for ``Computed Tomography'', ``MI'' for ``myocardial infarction'') should be considered a match.\\
    \quad \textbullet{} \textbf{Phrasing Variations:} The order of words or slight variations in phrasing should still be considered a match if the core components of the concept are present (e.g., ``aneurysm of the ascending aorta'' for ``aneurysm, ascending aorta'').\\
    \quad \textbullet{} \textbf{Substantial Representation:} The key elements of the concept must be present. A fleeting mention or an unrelated context does not count.\\
4. Finally, provide the total count of matched concepts as a single integer.\\
\\
\textbf{Example:}\\
\texttt{output\_string}: ``ct chest axial view showing a huge ascending aortic aneurysm (*).''\\
\texttt{concepts\_list}: ['x-ray computed tomography', 'aneurysm, ascending aorta', 'pulmonary embolism']\\
\\
\textbf{Reasoning Process:}\\
1. \texttt{x-ray computed tomography}: MATCH - ``ct'' in the output string is a common abbreviation for ``computed tomography''.\\
2. \texttt{aneurysm, ascending aorta}: MATCH - ``ascending aortic aneurysm'' in the output string directly refers to this concept.\\
3. \texttt{pulmonary embolism}: NO MATCH - There is no mention of pulmonary embolism or related terms in the output string.\\
\\
\textbf{Example Output Format:}\\
2\\
\\
---
\\
\textbf{Now, perform the task for the following inputs:}\\
\\
\texttt{output\_string}: \{output\}\\
\texttt{concepts\_list}: \{concepts\}\\
\\
\textbf{Provide the final count and brief reasoning process}\\
Count:\\
Reason:\\
\end{tcolorbox}

3. Prompt for Caption Prediction judge 

\begin{tcolorbox}[title=Prompt for Caption Prediction judge]
You are an expert in medical image caption analysis and semantic similarity. Your task is to score how closely the semantic meaning of a `Model Output` (a description of a medical image) matches a `Caption` (a reference description), by also considering a `Reference Concept List` extracted from the `Caption`.\\
\\
**Inputs:**\\
1.  `Caption`: The reference string describing the medical image.\\
2.  `Model Output`: The model-generated string describing the medical image.\\
3.  `Reference Concept List`: A list of key medical concepts explicitly mentioned or strongly implied in the `Caption`.\\
\\
**Scoring Rubric (0 to 3):**\\
\\
Your score should reflect how well the `Model Output` captures the essence of the `Caption`, with particular attention to whether the concepts in the `Reference Concept List` are appropriately represented in the `Model Output`.\\
\\
*   **Score 3 (High Semantic Correlation / Same Meaning):**\\
    *   The `Model Output` conveys essentially the same core medical information, findings, and context as the `Caption`.\\
    *   Most, if not all, concepts from the `Reference Concept List` are accurately and clearly represented in the `Model Output` (or their direct synonyms/equivalents).\\
    *   Minor differences in wording or specificity that do not alter the core meaning or the representation of reference concepts are acceptable.\\
\\
*   **Score 2 (Moderate Semantic Correlation):**\\
    *   The `Model Output` shares significant commonalities with the `Caption` and accurately represents a good portion (e.g., more than half) of the concepts from the `Reference Concept List`.\\
    *   Some reference concepts might be missing, generalized, or slightly misidentified in the `Model Output`, but the overall gist and several key concepts align.\\
    *   Alternatively, the `Model Output` might correctly represent the imaging modality and general anatomy but miss some key pathological concepts from the list or vice-versa.\\
\\
*   **Score 1 (Low Semantic Correlation):**\\
    *   The `Model Output` has only a superficial or very general overlap with the `Caption`.\\
    *   Only a few (or none, but with some general contextual similarity like modality) concepts from the `Reference Concept List` are identifiable in the `Model Output`, or they are represented inaccurately.\\
    *   The core medical message, guided by the reference concepts, is largely different.\\
\\
*   **Score 0 (No Semantic Correlation):**\\
    *   The `Model Output` describes entirely different findings, anatomical regions, or imaging modalities in a way that makes it irrelevant or contradictory to the `Caption`.\\
    *   Essentially none of the concepts from the `Reference Concept List` are found or accurately represented in the `Model Output`.\\
\\
\end{tcolorbox}
\begin{tcolorbox}[title=Prompt for Caption Prediction judge (Cont')]
**Evaluation Considerations (guided by the Reference Concept List):**\\
*   Does the `Model Output` reflect the correct imaging modality mentioned/implied by the concepts?\\
*   Does it identify the correct anatomical structures/regions from the concepts?\\
*   Does it describe the key findings/pathologies represented in the concepts?\\
*   Are qualifiers (e.g., location, number, severity) from the concepts maintained if critical?\\
\\
**Output Format:**\\
Return your score as an integer (0, 1, 2, or 3) and a brief `Reason` explaining your judgment based on the rubric, the `Caption`, the `Model Output`, and how well the `Reference Concept List` was reflected.\\
\\
Format your response exactly as follows:\\
Score: [0-3]\\
Reason: [Your brief explanation]\\
\\
**Examples:**\\
\\
**Example 1:**\\
`Caption:` anterior-posterior tibia-fibula radiographs of initial open tibia/fibula shaft fractures.\\
`Model Output:` ap x-ray of the left ankle showing a distal fibular fracture without significant displacement.\\
`Reference Concept List:` ['plain x-ray', 'lower extremity', 'postero-anterior']\\
Score: 1\\
Reason: `Model Output` matches 'plain x-ray' (ap x-ray) and 'lower extremity'. However, `Caption` refers to "tibia/fibula shaft fractures," while output specifies "distal fibular fracture" in the "left ankle," a more specific and potentially different location than implied by "shaft." Key concept of "tibia fracture" is missing. Limited alignment with core concepts.\\
\\
**Example 2:**\\
`Caption:` ct chest before starting chemotherapy showed bilateral parenchymal metastatic nodules(red arrow)\\
`Model Output:` axial cect thorax image showing a well-defined solitary pulmonary nodule in the left lower lobe (indicated by red arrow).\\
`Reference Concept List:` ['x-ray computed tomography', 'structure of parenchyma of lung', 'metastatic to', 'nodule']\\
Score: 2\\
Reason: `Model Output` matches 'x-ray computed tomography' (cect thorax), 'structure of parenchyma of lung' (pulmonary), and 'nodule'. However, `Caption` specifies "bilateral parenchymal metastatic nodules," while output describes a "solitary pulmonary nodule in the left lower lobe." The concepts 'metastatic to' and the plurality/bilaterality are missing or contradicted, which are key. Good overlap on modality and general pathology, but key characteristics differ.\\
\\
\end{tcolorbox}

\begin{tcolorbox}[title=Prompt for Caption Prediction judge (Cont')]
**Example 3:**\\
`Caption:` a ct scan of the chest. the scan shows a small sub-pleural nodule-like consolidation (white arrow).\\
`Model Output:` axial chest ct showing a peripheral, wedge-shaped area of consolidation in the left lower lobe, suggestive of pulmonary infarction (arrow).\\
`Reference Concept List:` ['x-ray computed tomography', 'nodule']\\
Score: 1\\
Reason: `Model Output` matches 'x-ray computed tomography' (chest ct). While both mention "consolidation," the `Caption`'s key finding is a "nodule-like consolidation." The `Model Output` describes a "wedge-shaped area of consolidation suggestive of pulmonary infarction," which is a different morphology and implication than a "nodule." The 'nodule' concept is poorly represented.\\
\\
**Example 4:**\\
`Caption:` chest radiograph during initial presentation demonstrating complete opacification of the right hemithorax with mediastinal shift to the opposite side.\\
`Model Output:` chest x-ray shows complete opacification of the left hemithorax with mediastinal shift to the left, consistent with left lung collapse.\\
`Reference Concept List:` ['plain x-ray', 'chest', 'postero-anterior', 'right thorax structure']\\
Score: 0\\
Reason: While `Model Output` matches 'plain x-ray' (chest x-ray) and 'chest', a critical concept 'right thorax structure' (opacification of *right* hemithorax) is directly contradicted by the output mentioning the *left* hemithorax. This fundamental difference in laterality makes the core medical information entirely different despite shared modality.\\
\\
---\\
\\
**Now, please score the following:**\\
\\
`Caption:` {caption}\\
`Model Output:` {model\_output}\\
`Reference Concept List:` {reference\_concept\_list}\\
\\
Score:\\
Reason:\\
\end{tcolorbox}

\newpage

\bibliography{sample}
\bibliographystyle{apalike}

\end{document}